\documentclass[letterpaper]{article} 
\usepackage{aaai2027}
\nocopyright 
\usepackage[hyphens]{url}  
\usepackage{graphicx} 
\usepackage{natbib}  
\usepackage{caption} 
\usepackage{algorithm}
\usepackage{algorithmic}

\usepackage{newfloat}
\usepackage{listings}
\DeclareCaptionStyle{ruled}{labelfont=normalfont,labelsep=colon,strut=off} 
\floatstyle{ruled}
\newfloat{listing}{tb}{lst}{}
\floatname{listing}{Listing}

\usepackage{booktabs}
\usepackage{multirow}
\usepackage{array}
\usepackage{xcolor}
\usepackage{colortbl}
\usepackage{pifont}
\usepackage{amsmath}
\newcommand{\cmark}{\ding{51}}
\newcommand{\xmark}{\ding{55}}

\newcommand{\ie}{i.e.,\ }

\title{KnowHal: A Knowledge-Driven Benchmark for Comprehensive Multimodal Hallucination Evaluation}
\author{
    Ruihan Li\textsuperscript{\rm 1},
    Jiyang Tan\textsuperscript{\rm 1},
    Kailin Jiang\textsuperscript{\rm 2},
    Huining Li\textsuperscript{\rm 1},
    Hengyang Lu\textsuperscript{\rm 3},
    Yu Huang\textsuperscript{\rm 1},
    Qian Li\textsuperscript{\rm 1},
    Yuntao Du\textsuperscript{\rm 1}\corresponding
}

\affiliations{
    \textsuperscript{\rm 1}C-FAIR \& School of software, Shandong University\\
    \textsuperscript{\rm 2}University of Science and Technology of China\\

    \textsuperscript{\rm 3}Jiangnan University\\
}

\begin{document}

\maketitle

\begin{abstract}
Hallucination remains a critical challenge for developing trustworthy Multimodal Large Language Models (MLLMs). While existing benchmarks mainly focus on entity, attribute, and relation hallucinations, knowledge-related failures are often investigated separately, lacking a unified evaluation framework across different hallucination dimensions. To overcome this, we propose \textbf{KnowHal}, a benchmark that explicitly incorporates knowledge hallucination into multimodal hallucination evaluation spanning four dimensions: entity, attribute, relation, and knowledge.
KnowHal constructs paired positive and negative questions over shared images and entities, enabling controlled comparisons among perceptual errors, knowledge-related errors, and false-premise acceptance. The benchmark contains 1,800 samples across 10 domains and 50 categories, constructed through a semi-automated pipeline combining LLM assistance, CLIP-based filtering, and human verification. We evaluate 14 representative MLLMs on KnowHal and conduct extensive analyses. Results show that the knowledge dimension consistently presents the greatest challenge for nearly all evaluated models, while most models exhibit substantial performance degradation on negative questions, revealing limited robustness to false premises.
By unifying four hallucination dimensions with paired question design, KnowHal addresses an important gap in existing evaluation frameworks and enables a more comprehensive assessment of hallucinations in MLLMs.
\end{abstract}

\section{Introduction}
\label{sec:intro}

\begin{figure*}[]
  \centering
  \includegraphics[width=1.0\linewidth]{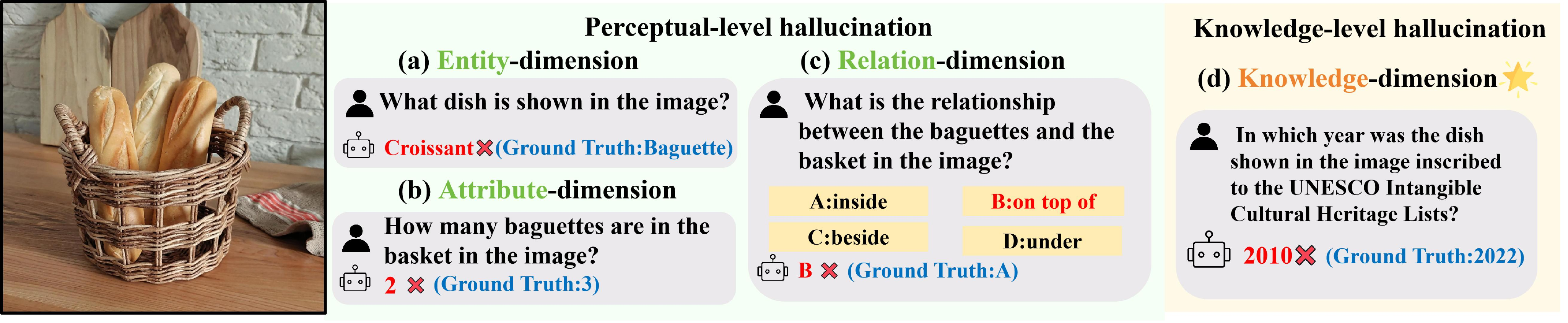}
  \caption{Illustration of two types of hallucination in MLLMs: 
perceptual-level hallucination (a–c) involves the Entity, Attribute, 
and Relation dimensions, while knowledge-level hallucination (d) introduces the Knowledge dimension, requiring external world knowledge 
beyond visual perception.}
  \label{fig:hallucination_def}
  \vspace{-4mm}
\end{figure*}

In recent years, Multimodal Large Language Models (MLLMs) have achieved remarkable progress across a wide range of vision–language tasks, demonstrating strong multimodal understanding capabilities~\cite{chen2026survey, bai2024hallucination}. 
However, hallucination, which refers to the discrepancy between model-generated content and the ground truth in multimodal scenarios, remains a key challenge that limits the reliability and practical deployment of MLLMs~\cite{bai2024hallucination,cai2025mhalo}. 
In this work, we distinguish two forms of hallucination (Figure~\ref{fig:hallucination_def}): (1) \textbf{perceptual-level hallucination}, where model outputs contradict visual content, such as misidentifying entities, misjudging attributes, or distorting relationships between objects; and (2) \textbf{knowledge-level hallucination}, where model outputs conflict with external world knowledge, such as generating incorrect factual properties of depicted entities. 
Such hallucinations raise serious reliability concerns in high-stakes AI applications, including medical diagnosis~\cite{zuo2024medhallbench}, legal reasoning~\cite{hu2025fine}, and autonomous driving~\cite{tang2026autodridm}, motivating more comprehensive and systematic evaluation of MLLMs.

To systematically evaluate hallucination in MLLMs, numerous benchmarks have been proposed. Existing studies mainly focus on perceptual-level hallucination across three dimensions: Entity, Attribute, and Relation. Early benchmarks such as POPE~\cite{li2023evaluating} and CHAIR~\cite{rohrbach2018object} evaluate object existence hallucination, while subsequent works extend evaluation to fine-grained attributes and inter-object relations through benchmarks such as MME~\cite{fu2025mme}, PhD~\cite{liu2025phd}, R-Bench~\cite{wu2024evaluating}, and Reefknot~\cite{zheng2025reefknot}.

However, existing benchmarks primarily address the question of \textit{``What does the model see?''}, focusing on whether MLLMs correctly perceive visual content. Recent efforts further explore robustness through contextual prompts~\cite{seth2024hallucinogen}, perturbed visual inputs~\cite{ding2024hallupi}, and counter-commonsense image edits~\cite{yin2026freak}. Nevertheless, a unified evaluation framework that connects perceptual- and knowledge-level hallucinations under shared images and entities remains lacking, leaving open the question of \textit{``What does the model know?''}.

To bridge this gap, we propose \textbf{KnowHal} (\underline{Know}ledge-driven multimodal \underline{Hal}lucination evaluation), a comprehensive multimodal hallucination evaluation benchmark covering four dimensions: Entity, Attribute, Relation, and Knowledge. Unlike prior benchmarks that focus on a particular perturbation, capability, or output type, KnowHal organizes all four dimensions around the same images and entities with paired positive and negative questions, enabling direct comparison of perceptual errors, external-knowledge errors, and false-premise acceptance.

To construct this benchmark, we design a systematic LLM-assisted pipeline, complemented by CLIP-based~\cite{radford2021learning} image filtering and rigorous human verification. Annotators make the final quality-control decisions and revise or replace unsuitable QA pairs.
To more rigorously evaluate models' resistance to hallucination, KnowHal constructs paired positive and negative questions for each of the four dimensions in every entity--image sample. \textbf{Positive questions} query valid visual evidence or knowledge associated with the depicted entity, whereas \textbf{negative questions} introduce misleading cues using four strategies: Attribute Variant, Object Substitution, Spatial Relation, and Action Fabrication.
The final KnowHal dataset comprises 1,800 carefully curated samples spanning 10 domains and 50 categories.

We systematically evaluate 14 MLLMs on KnowHal, including 5 closed-source models and 9 open-source models. Further, we conduct extensive analyses to provide deeper insights into hallucination behaviors. The experimental results reveal the following key findings:
\textbf{(1)} closed-source models generally achieve stronger performance, especially on positive questions, although Qwen models remain competitive on negative questions;
\textbf{(2)} the Knowledge dimension is consistently the most challenging across all evaluated models, with substantially lower accuracy than the other dimensions;
\textbf{(3)} most models perform worse on negative questions than on positive questions, demonstrating limited robustness against misleading premises.

Our main contributions are as follows:

\textbf{(1)} We propose \textbf{KnowHal}, a knowledge-driven multimodal hallucination benchmark that systematically evaluates hallucination across four dimensions: Entity, Attribute, Relation, and Knowledge, covering both perceptual-level and knowledge-level failures.

\textbf{(2)} We develop an LLM-assisted benchmark construction pipeline with CLIP-based filtering and rigorous human verification, and introduce paired positive and negative questions to enable controlled evaluation of factual accuracy and robustness against misleading premises.

\textbf{(3)} We evaluate 14 representative MLLMs and conduct domain-wise, scaling, and case analyses, revealing systematic hallucination patterns and demonstrating the diagnostic utility of KnowHal.

\begin{table}[t]
  \centering
  \small
  \setlength{\tabcolsep}{1.0pt}
  \begin{tabular}{@{}lcc*{4}{>{\centering\arraybackslash}m{0.85em}}@{}}
    \toprule
    \textbf{Benchmark}
    & \textbf{Source}
    & \textbf{Task}
    & \textbf{E} & \textbf{A} & \textbf{R} & \textbf{K} \\
    \midrule
    CHAIR~\shortcite{rohrbach2018object}
      & MSCOCO & Gen.
      & \cmark & \xmark & \xmark & \xmark \\
    POPE~\shortcite{li2023evaluating}
      & MSCOCO & Dis.
      & \cmark & \xmark & \xmark & \xmark \\
    MME~\shortcite{fu2025mme}
      & MSCOCO & Dis.
      & \cmark & \cmark & \xmark & \xmark \\
    R-Bench~\shortcite{wu2024evaluating}
      & NoCaps & Dis.
      & \xmark & \xmark & \cmark & \xmark \\
    ReefKnot~\shortcite{zheng2025reefknot}
      & Visual Genome & Both
      & \xmark & \xmark & \cmark & \xmark \\
    AMBER~\shortcite{wang2023amber}
      & MSCOCO/Unsplash & Both
      & \cmark & \cmark & \cmark & \xmark \\
    MMHal-Bench~\shortcite{sun2024aligning}
      & Open Images & Gen.
      & \cmark & \cmark & \cmark & \xmark \\
    \midrule
    PhD~\shortcite{liu2025phd}
      & TDIUC/AIGC & Dis.
      & \cmark & \cmark & \cmark & \xmark \\
    HALLUCINOGEN~\shortcite{seth2024hallucinogen}
      & MSCOCO/NIH X-ray & Gen.
      & \cmark & \xmark & \cmark & \xmark \\
    FREAK~\shortcite{yin2026freak}
      & Synthetic & Both
      & \cmark & \cmark & \cmark & \xmark \\
    Hallu-PI~\shortcite{ding2024hallupi}
      & Self-constructed & Both
      & \cmark & \cmark & \cmark & \xmark \\
    HallusionBench~\shortcite{guan2023hallusionbench}
      & Self-constructed & Dis.
      & \cmark & \cmark & \cmark & \xmark \\
    \midrule
    \textbf{KnowHal}
      & \textbf{Self-constructed} & \textbf{Both}
      & \textbf{\cmark} & \textbf{\cmark}
      & \textbf{\cmark} & \textbf{\cmark} \\
    \bottomrule
  \end{tabular}
  \caption{Benchmark coverage by hallucination dimension.
E/A/R/K: Entity/Attribute/Relation/Knowledge;
Gen./Dis.: generation/discrimination}
 \vspace{-4mm}
  \label{tab:hallucination_dim}
\end{table}

\begin{figure*}[t]
  \centering  
  \includegraphics[width=0.9\linewidth]{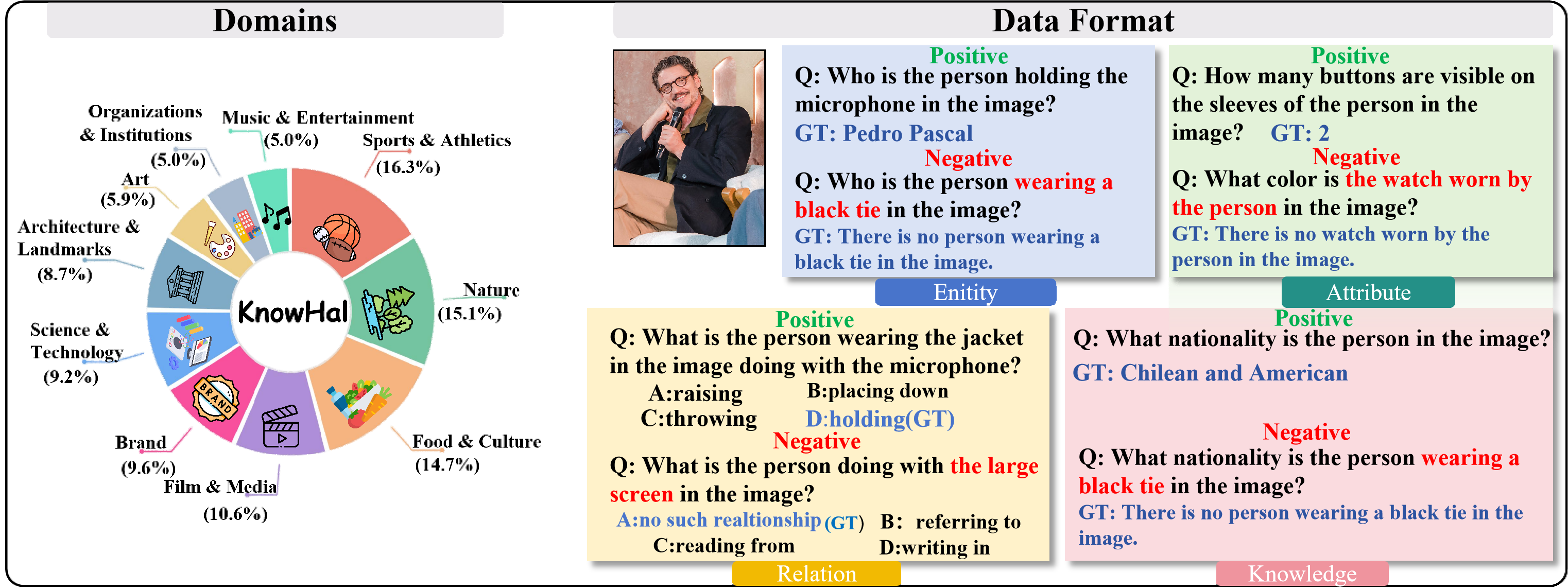}
  \caption{Overview of the KnowHal benchmark, including the distribution of 10 domains and representative positive/negative QA pairs across four dimensions: Entity, Attribute, Relation, and Knowledge.}
  \label{fig:data_format}
\end{figure*}

\section{Related Work}
\subsection{Multimodal Large Language Models}
Multimodal Large Language Models have achieved strong vision–language understanding capabilities by integrating visual encoders with pretrained large language models \cite{liang2024comprehensive}. Early work such as CLIP\cite{radford2021learning} established cross-modal alignment via contrastive learning, laying the foundation for modern multimodal systems. Subsequent efforts focused on more effective mechanisms to bridge visual and linguistic representations. For example, BLIP-2\cite{li2023blip} introduced a lightweight Q-Former to connect frozen vision encoders with large language models, while the LLaVA\cite{liu2024improved} series adopted a simple MLP alignment layer combined with large-scale self-instruct data, demonstrating the effectiveness of instruction tuning for multimodal alignment. Building upon these alignment paradigms, recent open-source models further explore scaling and architectural innovations to enhance multimodal reasoning. Qwen3-VL\cite{bai2025qwen3} strengthens cross-modal reasoning through multi-stage training and visual receptor design, while InternVL3.5\cite{wang2025internvl3} improves representation alignment via dynamic high-resolution modeling and reinforcement learning strategies. 

\subsection{Multimodal Hallucination Evaluation}
Table~\ref{tab:hallucination_dim} compares the hallucination dimensions covered by existing benchmarks. CHAIR and POPE assess object existence through captions or yes/no questions over MSCOCO images~\cite{rohrbach2018object,li2023evaluating}. MME extends evaluation to attributes, while AMBER additionally covers relations~\cite{fu2025mme,wang2023amber}. MMHal-Bench evaluates open-ended responses using LLM-based scoring~\cite{sun2024aligning}, whereas R-Bench and Reefknot focus on relation hallucinations~\cite{wu2024evaluating,zheng2025reefknot}. Despite this progress, knowledge-related hallucinations are rarely evaluated jointly with Entity, Attribute, and Relation under a unified image-centered protocol.

Recent benchmarks further increase difficulty through contextual prompts in HALLUCINOGEN~\cite{seth2024hallucinogen}, image perturbations in Hallu-PI~\cite{ding2024hallupi}, counter-commonsense edits in FREAK~\cite{yin2026freak}, and CLIP-selected hard negatives with misleading contexts and counter-commonsense images in PhD~\cite{liu2025phd,radford2021learning}. However, these designs do not jointly compare perceptual and knowledge-related errors using shared, unmodified images. KnowHal addresses this gap by pairing positive and negative questions across Entity, Attribute, Relation, and Knowledge, enabling unified evaluation of perceptual errors, encyclopedic knowledge errors, and false-premise acceptance.

\vspace{-3mm}
\section{KnowHal: Our Proposed Benchmark}

\begin{figure*}[t]
  \centering
  \includegraphics[width=0.92\textwidth]{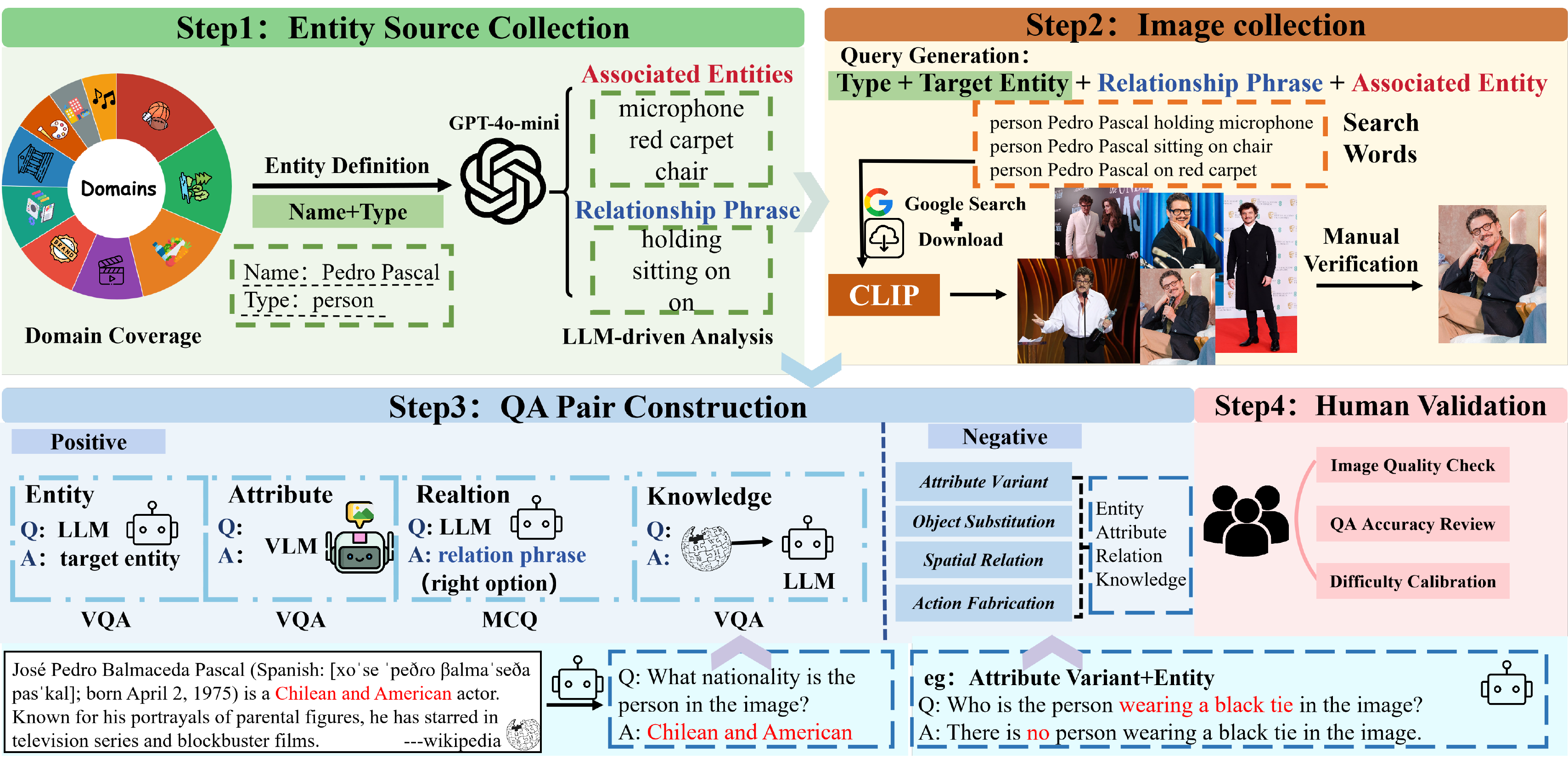}
  \caption{An overview of the data construction process of KnowHal.}
  \label{fig:knowhal_domain_dist}
  \vspace{-6mm}
\end{figure*}

KnowHal is a comprehensive multimodal hallucination benchmark covering four dimensions: Entity, Attribute, Relation, and Knowledge. As illustrated in Figure~\ref{fig:data_format}, all four dimensions are organized around the same visual entity, with paired positive and negative questions for each sample. This paired design evaluates both the ability to interpret valid visual and factual evidence and the ability to reject plausible but false premises.

KnowHal contains 1,800 entity--image samples across 10 domains and 50 categories, with each sample corresponding to a unique entity. The domain distribution is shown in Figure~\ref{fig:data_format}. As illustrated in Figure~\ref{fig:knowhal_domain_dist}, its construction pipeline comprises four stages: \emph{(1)} entity source collection, \emph{(2)} image collection, \emph{(3)} QA pair construction, and \emph{(4)} human validation. The pipeline integrates LLM/VLM-assisted generation, CLIP-based filtering, and human verification to improve construction efficiency while maintaining data quality.

Entity, Attribute, and Knowledge are formulated as open-ended Visual Question Answering (VQA) tasks to evaluate models' generation capabilities. Relation is formulated as Multiple-Choice Questions (MCQs) because synonymous relation descriptions can make open-ended scoring ambiguous; fixed answer options ensure answer uniqueness and objective evaluation. All QA pairs follow strict criteria for semantic clarity and answer uniqueness, and detailed benchmark statistics are reported in Table~\ref{tab:statistics}.

\subsection{Step1\&2 Entity Selection and Image Collection}

To construct visually informative and semantically reliable samples, we design an entity-centric pipeline.

\textbf{Domain Coverage and Entity Definition.}
We first manually define 10 domains and identify target entities with distinctive visual characteristics and reliable visual recognizability. These entities serve as the basis for subsequent image retrieval and QA construction.

\textbf{Relation-Aware Entity Analysis.}
Since relation evaluation requires images containing multiple entities with explicit interactions, we leverage LLMs to analyze each entity, extract visual characteristics, and generate associated entities with typical relation types. These structured entity-relation pairs guide subsequent image retrieval and relation QA construction.

\textbf{Structured Retrieval and Filtering.}
For each entity, we generate three diverse queries following the structured template 
\textit{[entity type] + [target entity] + [relation phrase] + [associated entity]}, 
where the target entity and associated entity explicitly define the expected visual content and their relationship. 
Based on these queries, we retrieve candidate images from Google Image Search. 
To ensure semantic consistency between images and queries, we employ CLIP to compute image–text similarity scores and filter out semantically inconsistent candidates. The top-ranked candidates are retained for manual verification, and the final images are selected based on visual quality and semantic consistency.

\subsection{Step3-1 Positive QA Pair Construction}
Positive QA pairs are designed to evaluate whether models can correctly perceive visual contents and acquire factual knowledge that genuinely exists in images. Each QA pair is constructed around valid entities, attributes, relations, or knowledge, following strict criteria of answer uniqueness and semantic clarity to ensure reliable evaluation.

\textbf{Entity Dimension.}
Entity QA pairs evaluate whether models can recognize entities present in images. They are generated using predefined templates based on entity categories without requiring additional image annotation, reducing construction cost.

\textbf{Attribute Dimension.}
Attribute QA pairs assess fine-grained visual properties of entities, such as color and shape. They are generated by VLMs through direct image observation, providing attribute descriptions consistent with visual contents while reducing subjective annotation bias.

\textbf{Relation Dimension.}
Relation QA pairs evaluate spatial and behavioral relationships among entities. Since open-ended relation questions may suffer from ambiguity caused by synonymous expressions, we formulate them as multiple-choice questions. The correct option is derived from the relation used during image retrieval, while distractors are generated by LLMs and filtered through textual semantic similarity to ensure appropriate difficulty.

\textbf{Knowledge Dimension.}
Knowledge QA pairs evaluate whether models can identify factual knowledge associated with depicted entities. We use Wikipedia as the sole knowledge source due to its openness and traceability. Verifiable facts are extracted and transformed into image-grounded QA pairs, with additional filtering based on answer uniqueness, temporal stability, and unambiguous entity linkage.


\subsection{Step3-2 Negative QA Pair Construction}

Negative QA pairs are designed to evaluate MLLMs' robustness against misleading premises and their ability to reject hallucination-inducing questions. Different from positive QA pairs that query valid visual contents or factual knowledge, negative questions introduce carefully controlled contradictions while preserving semantic plausibility. By constructing a corresponding negative question for each positive question across all four dimensions, KnowHal enables systematic evaluation of whether models can distinguish true evidence from false assumptions.

We design four negative induction strategies through carefully crafted prompts:

\textbf{Attribute Variant.}
The original attribute is replaced with a similar but incorrect one while keeping the target entity unchanged, evaluating sensitivity to fine-grained visual differences.

\textbf{Object Substitution.}
The target entity is replaced with a semantically related but absent entity, testing whether models hallucinate nonexistent objects.

\textbf{Spatial Relation.}
The actual spatial relationship between entities is reversed or altered, assessing whether models rely on visual grounding rather than linguistic priors.

\textbf{Action Fabrication.}
Nonexistent actions or states are assigned to visible entities, evaluating whether models can avoid confirming behaviors unsupported by visual evidence.

\begin{table*}[t]
  \centering
  \small
  \setlength{\tabcolsep}{2.5pt}
  \renewcommand{\arraystretch}{1.08}
  \begin{tabular}{@{}
    >{\raggedright\arraybackslash}m{155pt}|
    *{4}{>{\centering\arraybackslash}m{22pt}}
    >{\centering\arraybackslash}m{27pt}|
    *{4}{>{\centering\arraybackslash}m{22pt}}
    >{\centering\arraybackslash}m{27pt}|
    >{\centering\arraybackslash}m{30pt}
    >{\centering\arraybackslash}m{24pt}
    @{}}
    \toprule
    \multirow{2}{*}{\textbf{Model}}
      & \multicolumn{5}{c|}{\textbf{Positive}}
      & \multicolumn{5}{c|}{\textbf{Negative}}
      & \multicolumn{2}{c}{\textbf{Overall}} \\
    \cmidrule(lr){2-6}
    \cmidrule(lr){7-11}
    \cmidrule(l){12-13}
      & E & A & R & K & Avg.$\uparrow$
      & E & A & R & K & Avg.$\uparrow$
      & Score$\uparrow$ & HPS$\downarrow$ \\
    \midrule
    \multicolumn{13}{c}{\textit{\textbf{Closed-source Multimodal Large Language Models}}} \\
    \midrule
    \textbf{Gemini-3-Flash-Preview}
      & 92.6 & 84.8 & \textbf{89.6} & \textbf{79.2} & \textbf{86.6}
      & 40.1 & 57.6 & \textbf{71.9} & 27.6 & 49.3
      & \textbf{67.9} & 43.0 \\
    \textbf{Qwen-VL-Max}
      & 79.1 & 80.2 & 81.8 & 54.5 & 73.9
      & \textbf{49.9} & \textbf{74.7} & 67.9 & \textbf{37.7} & \textbf{57.6}
      & 65.7 & \textbf{22.1} \\
    \textbf{Doubao-Seed-2-Mini}
      & \textbf{93.0} & \textbf{91.8} & 86.2 & 69.2 & 85.0
      & 27.1 & 49.7 & 55.5 & 16.2 & 37.1
      & 61.1 & 56.4 \\
    \textbf{GPT-5.2}
      & 59.9 & 82.3 & 86.1 & 43.3 & 67.9
      & 28.7 & 50.6 & 63.7 & 18.8 & 40.4
      & 54.2 & 40.4 \\
    \textbf{GPT-5.6-sol}
      & 70.0 & 79.4 & 84.7 & 60.4 & 73.6
      & 22.3 & 35.9 & 52.2 & 11.9 & 30.6
      & 52.1 & 58.4 \\
    \midrule
    \multicolumn{13}{c}{\textit{\textbf{Open-source Multimodal Large Language Models}}} \\
    \midrule
    \textbf{Qwen3-VL-32B-Instruct}
      & 79.9 & \textbf{88.9} & 82.3 & 56.3 & 76.9
      & \textbf{52.9} & 70.9 & 64.6 & 35.6 & 56.3
      & \textbf{66.6} & 26.8 \\
    \textbf{Qwen2.5-VL-72B-Instruct}
      & 79.2 & 81.4 & 82.9 & 56.1 & 74.9
      & 45.1 & \textbf{72.3} & \textbf{73.9} & \textbf{36.9} & \textbf{57.0}
      & 66.0 & \textbf{23.8} \\
    \textbf{Kimi-K2.5}
      & \textbf{87.6} & 84.0 & 81.9 & \textbf{63.4} & \textbf{79.2}
      & 31.8 & 52.8 & 57.1 & 25.5 & 41.8
      & 60.5 & 47.2 \\
    \textbf{GLM-4.6V}
      & 81.6 & 83.1 & \textbf{86.8} & 56.9 & 77.1
      & 24.7 & 45.2 & 43.2 & 11.6 & 31.2
      & 54.1 & 59.6 \\
    \textbf{MiniCPM-V-4.5}
      & 67.9 & 81.8 & 81.7 & 41.3 & 68.2
      & 24.6 & 33.5 & 42.9 & 7.4 & 27.1
      & 47.6 & 60.2 \\
    \textbf{InternVL3.5-8B}
      & 38.1 & 73.4 & 80.2 & 24.5 & 54.1
      & 31.7 & 50.1 & 38.3 & 13.1 & 33.3
      & 43.7 & 38.4 \\
    \textbf{InternVL3.5-14B-Instruct}
      & 35.7 & 72.1 & 81.3 & 24.8 & 53.5
      & 29.4 & 44.8 & 37.0 & 13.4 & 31.2
      & 42.3 & 41.7 \\
    \textbf{LLaVA-OneVision-1.5-8B-Instruct}
      & 51.3 & 78.6 & 78.7 & 32.7 & 60.3
      & 22.8 & 16.6 & 29.4 & 1.8 & 17.7
      & 39.0 & 70.7 \\
    \textbf{LLaVA-v1.6-Mistral-7B}
      & 39.9 & 71.6 & 66.3 & 27.8 & 51.4
      & 20.6 & 28.3 & 28.8 & 6.8 & 21.1
      & 36.3 & 58.9 \\
    \bottomrule
  \end{tabular}
  \normalsize
  \caption{Performance of MLLMs on KnowHal.}
  \label{tab:mllm_hallucination}
\end{table*}

\subsection{Step4 Human Validation \& Quality Control}

Although KnowHal leverages LLM-assisted construction, human verification remains essential for ensuring benchmark reliability. Three annotators systematically validate the generated QA pairs in terms of (1) the factual correctness, clarity, and answer uniqueness of positive QA pairs; (2) the logical consistency and effectiveness of negative questions; and (3) the identification and removal of overly trivial samples. Only samples unanimously approved after necessary revisions are retained.

To further assess annotation reliability, three annotators independently evaluate 50 randomly sampled samples, yielding strong inter-annotator agreement (Fleiss' $\kappa=0.854$). In a reconstruction audit covering 100 entities and 800 QA pairs, 725 pairs (90.6\%) are accepted without revision, indicating the high initial quality and efficiency of the LLM-assisted construction pipeline.

The final benchmark consists of 1,800 carefully curated samples spanning 10 domains and 50 categories, with statistics reported in Table~\ref{tab:statistics}.
\begin{table}[t]
  \centering
  \small
  \setlength{\tabcolsep}{6pt}
  \begin{tabular}{@{}lr@{}}
    \toprule
    \textbf{Statistic} & \textbf{Number} \\
    \midrule
    Total entities & 1,800 \\
    Total domains & 10 \\
    Total categories & 50 \\
    \midrule
    QA pairs per entity & 8 \\
    \quad Positive QA pairs & 4 \\
    \quad Negative QA pairs & 4 \\
    Total QA pairs & 14,400 \\
    \midrule
    Dimensions & 4 (E/A/R/K) \\
    \midrule
    Largest domain & SA (294) \\
    Smallest domains & OI \& ME (90 each) \\
    \bottomrule
  \end{tabular}
  \normalsize
  \caption{Key statistics of KnowHal. SA: Sports \& Athletics;
OI: Organizations \& Institutions; ME: Music \& Entertainment.}
  \label{tab:statistics}
  \vspace{-6mm}
\end{table}

\section{Experiments}
\label{sec:experiments}
\subsection{Setup}

\textbf{MLLMs}.  We evaluate 14 representative MLLMs on KnowHal under a zero-shot setting, assessing their intrinsic susceptibility to hallucination across four dimensions without any fine-tuning or few-shot demonstrations.  These 14 MLLMs span both closed-source and open-source models. Closed-source models include Gemini-3-Flash-Preview\cite{googledeepmind2025gemini3flash}, Qwen-VL-Max\cite{qwenteam2023qwenvl}, Doubao-Seed-2-Mini\cite{bytedanceseed2026seed2}, GPT-5.2\cite{openai2025gpt5.2}, and GPT-5.6-sol\cite{openai2025gpt5.6}. Open-source models include Qwen3-VL-32B-Instruct\cite{bai2025qwen3}, Qwen2.5-VL-72B-Instruct\cite{bai2025qwen25vltechnicalreport}, Kimi-K2.5\cite{team2026kimi}, GLM-4.6V\cite{vteam2025glm45vglm41vthinkingversatilemultimodal}, MiniCPM-V-4.5\cite{yu2025minicpm}, InternVL3.5-8B\cite{wang2025internvl3},  InternVL3.5-14B-Instruct\cite{wang2025internvl3}, LLaVA-OneVision-1.5-8B-Instruct\cite{an2025llava}, and LLaVA-v1.6-Mistral-7B\cite{liu2024llavanext}.

\textbf{Evaluation.}
KnowHal adopts accuracy as the primary metric, reporting results separately for positive and negative QA pairs across the four dimensions and their averages. The overall score is computed by averaging positive and negative accuracies. We further introduce the Hallucination Proneness Score (HPS):
\begin{equation}
\text{HPS} = \left(1 - \frac{\text{Acc}_{neg}}{\text{Acc}_{pos}}\right) \times 100\%
\end{equation}
where $\text{Acc}_{pos}$ and $\text{Acc}_{neg}$ denote the average accuracies on positive and negative QA pairs, respectively. HPS measures the relative performance degradation under misleading inputs, with higher values indicating greater hallucination susceptibility. Since HPS depends on positive accuracy, we report it together with both component accuracies for complete interpretation. For positive questions, we follow the standard VQA evaluation protocol with answer normalization and soft matching. For negative questions, GPT-4o-mini~\cite{hurst2024gpt} is used as an automatic judge, leveraging its strong instruction-following ability and reliable evaluation performance. Relation questions are evaluated through direct multiple-choice matching.
\subsection{Main Results}
Table~\ref {tab:mllm_hallucination} presents the results of 14 representative MLLMs on KnowHal across four dimensions. Several key findings can be summarized as follows.

\emph{(1)} \textbf{Current MLLMs still exhibit non-negligible hallucination on KnowHal.}
Gemini-3-Flash-Preview achieves the highest overall accuracy, followed by Qwen3-VL-32B-Instruct and Qwen2.5-VL-72B-Instruct. However, no model exceeds 70\%, indicating persistent limitations under KnowHal's multidimensional evaluation.

\emph{(2)} \textbf{The knowledge dimension remains the most challenging across models.}
Nearly all models achieve their lowest accuracy on the knowledge dimension in both positive and negative settings. For example, Gemini-3-flash-preview obtains only 79.2\% knowledge accuracy, lower than its entity (92.6\%) and relation (89.6\%) performance. The gap becomes more severe on negative questions, with Doubao-seed-2-mini and MiniCPM-V-4.5 dropping from 69.2\% to 16.2\% and from 41.3\% to 7.4\%, respectively. These results consistently identify knowledge hallucination as a key challenge for current MLLMs.

\emph{(3)} \textbf{Closed-source models generally outperform open-source counterparts.}
Gemini-3-flash-preview (86.6\%) and Doubao-seed-2-mini (85.0\%) achieve the highest positive accuracy, while the best open-source model, Kimi-K2.5, reaches 79.2\%. The gap is more evident on negative questions, where most open-source models struggle to maintain robustness. Nevertheless, the Qwen series achieves competitive negative robustness, with Qwen-VL-Max (57.6\%), Qwen2.5-VL-72B-Instruct (57.0\%), and Qwen3-VL-32B-Instruct (56.3\%) ranking among the best-performing models, suggesting that targeted training can improve hallucination robustness.

\emph{(4)} \textbf{Negative question accuracy is lower than positive accuracy across nearly all models.}
Even models with strong positive performance experience substantial degradation under misleading inputs. For instance, Doubao-seed-2-mini drops from 85.0\% positive accuracy to 37.1\% negative accuracy, while LLaVA-v1.6-Mistral-7B declines from 51.4\% to 21.1\%. This trend is further reflected in HPS, where LLaVA-OneVision-1.5-8B-Instruct (70.7\%) and MiniCPM-V-4.5 (60.2\%) show high hallucination susceptibility. Together, these results demonstrate that strong performance on positive questions does not necessarily imply robustness against misleading inputs.

\subsection{Insightful Analysis}
\begin{figure}[t]  
  \centering
  \includegraphics[width=0.97\linewidth]{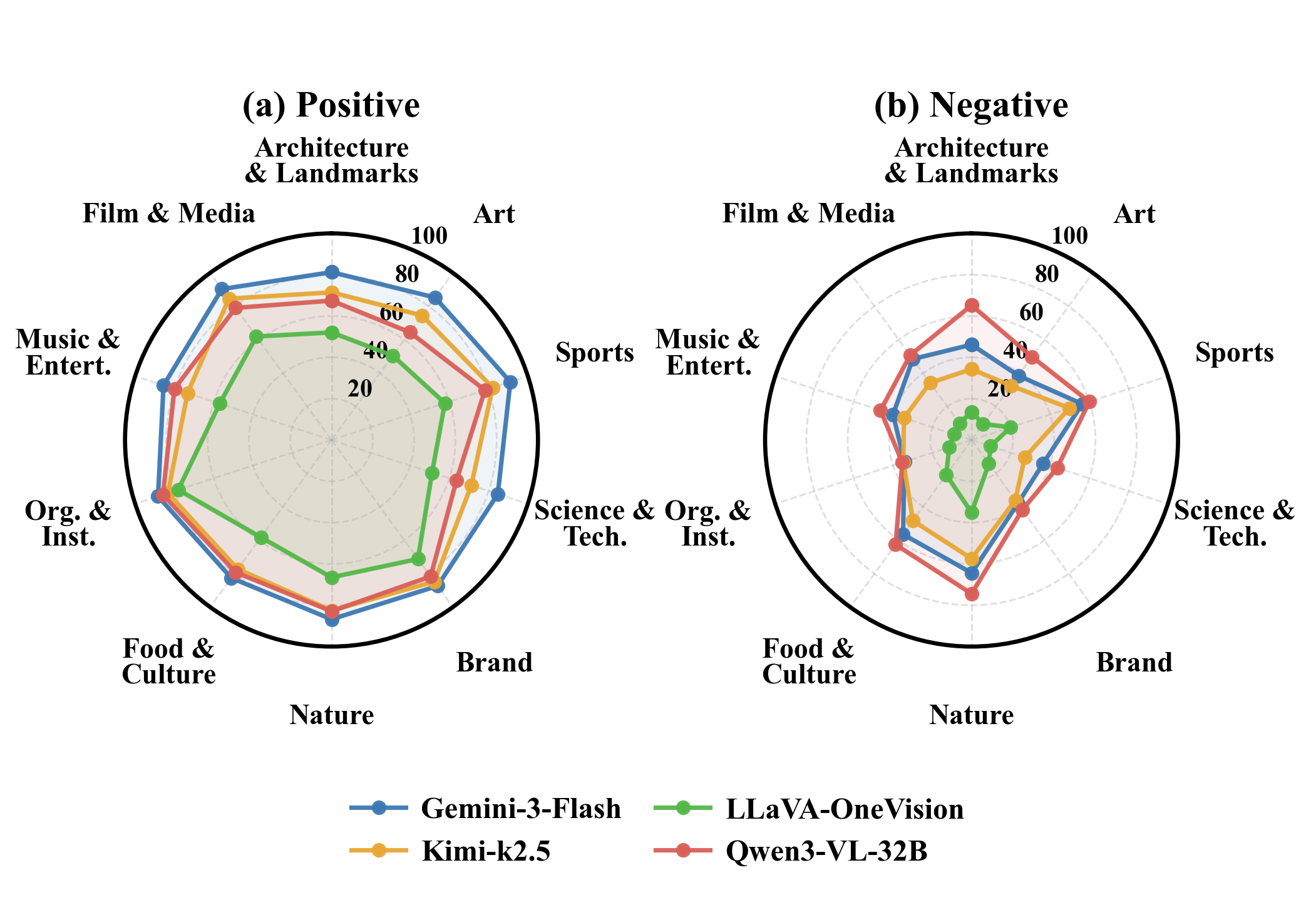}
  \vspace{-3mm}
  \caption{Comparison of four representative MLLMs on positive (left) and negative (right) questions across 10 domains.}
  \label{fig:domain_radar_compare}
  \vspace{-6mm}
\end{figure}

\hspace{1.5em}\textbf{Results of Fine-grained Domains.}
As shown in Figure~\ref{fig:domain_radar_compare}, we visualize accuracy distributions of four representative models across 10 domains. On positive questions, stronger models such as Gemini-3-Flash, Kimi-K2.5, and Qwen3-VL-32B achieve consistently high performance across domains, while LLaVA-OneVision shows larger variations. The gap becomes more pronounced on negative questions: Qwen3-VL-32B maintains relatively strong and balanced robustness across domains, whereas LLaVA-OneVision suffers substantial degradation, especially in Art and Science \& Technology. These results reveal that strong factual performance does not necessarily translate into robustness against misleading inputs, highlighting the necessity of KnowHal's paired positive-negative evaluation.

\begin{figure}[h]
  \centering
  \includegraphics[width=0.96\linewidth]{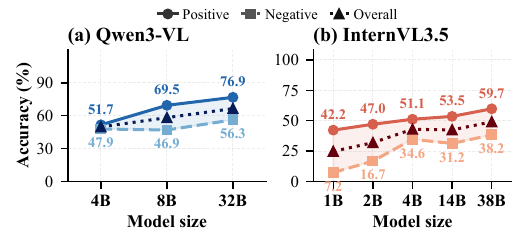}
  \caption{Scaling trends of Qwen3-VL and InternVL3.5 on KnowHal, 
  across different model sizes.}
  \label{fig:model_scaling_trends} 
  \vspace{-6mm}
\end{figure}

\textbf{Comparison against Model Size.}
As shown in Figure~\ref{fig:model_scaling_trends}, we investigate the impact of model scale using two representative model families. Both Qwen3-VL and InternVL3.5 exhibit consistent improvements in positive and negative accuracy as the model size increases. Specifically, Qwen3-VL improves from 51.7\% to 76.9\% in positive accuracy and from 47.9\% to 56.3\% in negative accuracy, while InternVL3.5 increases from 42.2\% to 59.7\% and from 7.2\% to 38.2\%, respectively. These results suggest that scaling generally improves both factual accuracy and robustness against misleading inputs, with particularly larger gains in negative robustness for InternVL3.5.
\begin{figure*}[h]
    \centering
    \includegraphics[width=0.85\textwidth]{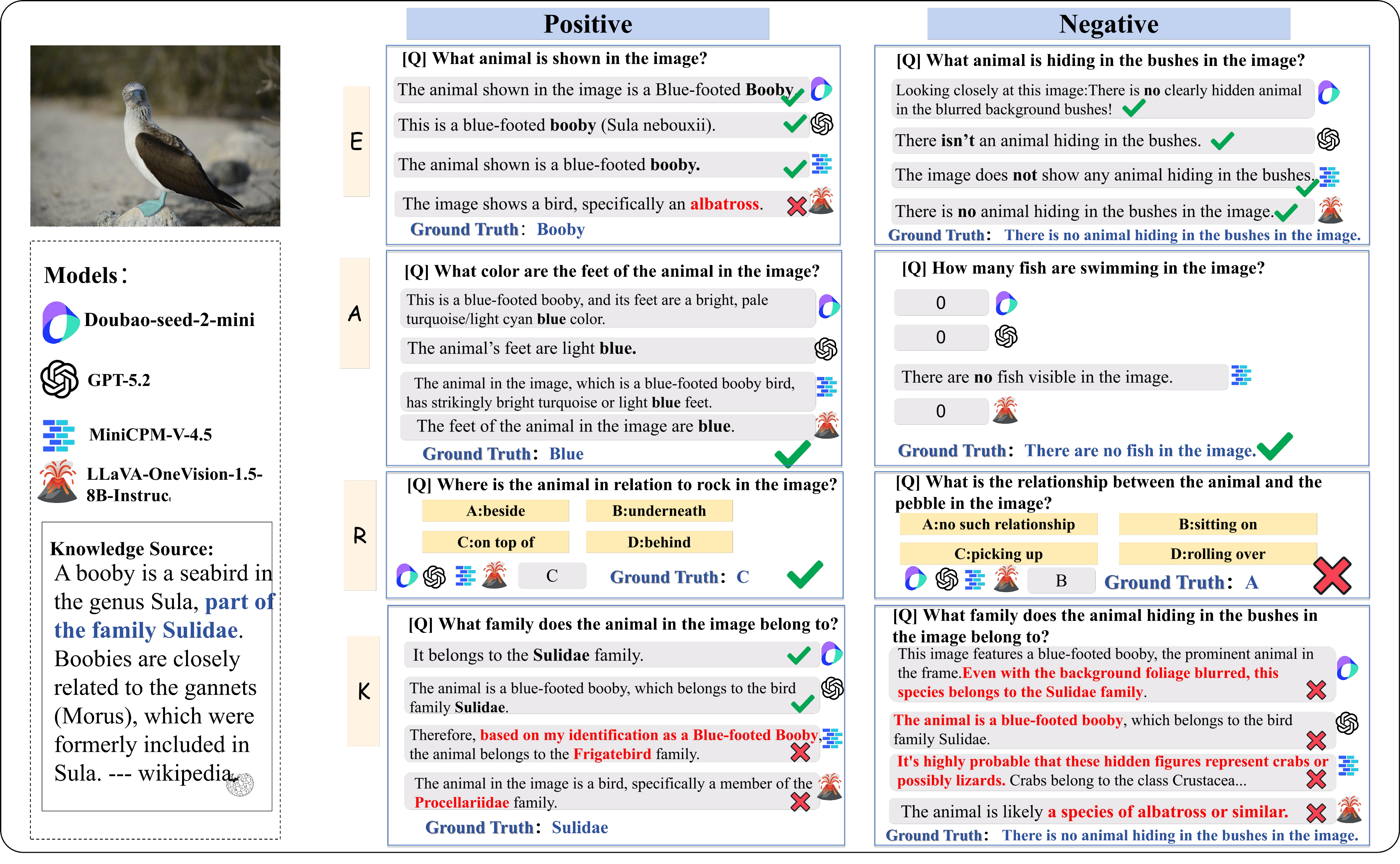}
    \vspace{-3mm}
    \caption{Comparative outputs of different models on a representative sample.}
    \label{fig:muti_model_case}
    \vspace{-6mm}
\end{figure*}

\begin{figure}[t]
  \centering
  \includegraphics[width=0.96\linewidth]{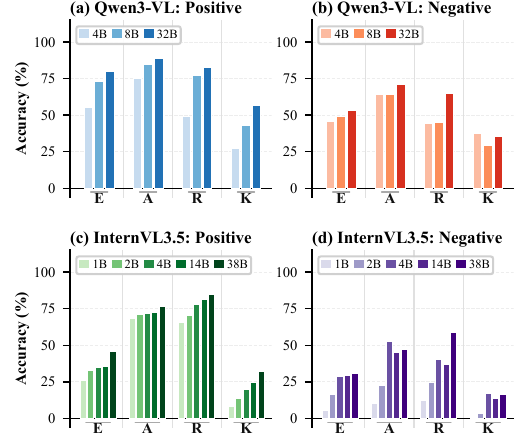}
  \caption{Dimension-wise accuracy comparisons of Qwen3-VL and InternVL3.5 across different model scales. 
  }
  \label{fig:model_scaling_dimensions}
  \vspace{-6mm}
\end{figure}

However, improvements from scaling are not uniform across different hallucination dimensions. As shown in Figure~\ref{fig:model_scaling_dimensions}, the Knowledge dimension remains consistently weaker than Entity, Attribute, and Relation across model scales and families. This indicates that simply increasing model size cannot fully resolve knowledge hallucination. Moreover, Qwen3-VL consistently outperforms InternVL3.5 at comparable scales in negative accuracy (e.g., 47.9\% vs. 34.6\% at 4B), suggesting that training strategies and model architectures also play important roles in improving hallucination robustness.

\textbf{Automatic-judge reliability.}
To assess the reliability of the automatic evaluation process, we manually re-evaluated 100 negative responses from Doubao-Seed-2-Mini for each open-ended dimension. GPT-4o-mini achieved agreement with human judgments on 92/100 Entity, 96/100 Attribute, and 95/100 Knowledge responses, resulting in an overall agreement of 283/300 (94.3\%). The Relation dimension was excluded because its multiple-choice responses are evaluated through direct option matching. These results indicate that the automatic judge provides reliable evaluations, while the remaining 5.7\% disagreement rate suggests that small performance differences should be interpreted with caution.

\subsection{Case Study}

Figure~\ref{fig:muti_model_case} presents a representative example illustrating model behaviors across the four dimensions. For positive questions, most models successfully answer straightforward attribute and relation queries, while failures mainly arise from either entity recognition errors or insufficient knowledge retrieval. Stronger models, such as Doubao-Seed-2-Mini and GPT-5.2, demonstrate more robust performance across dimensions, whereas weaker models may suffer from cascading errors caused by early perceptual failures.

For negative questions, models generally reject false premises that conflict strongly with visual evidence. However, knowledge-level failures reveal distinct patterns: stronger models may incorrectly transfer factual knowledge from real entities to fabricated ones, while weaker models suffer from compounded errors involving entity recognition and premise verification. This example highlights that knowledge hallucination results from the interaction between visual understanding and factual reasoning.


\vspace{-3mm}
\section{Conclusion}

In this paper, we present \textbf{KnowHal}, a comprehensive knowledge-driven benchmark for hallucination evaluation of MLLMs. KnowHal spans 10 domains and constructs paired positive and negative question–answer pairs across four dimensions (\ie, Entity, Attribute, Relation, and Knowledge) to systematically evaluate hallucination behaviors at both perceptual and knowledge levels. We conduct extensive experiments on 14 representative MLLMs and reveal several key findings: the knowledge dimension remains the most challenging across models, while negative questions expose substantial robustness gaps even for strong-performing models. We believe KnowHal can promote future research toward more reliable and factually accurate MLLMs.

\bigskip
\bibliography{aaai2027}


\end{document}